\documentclass[9pt]{article}
\usepackage{spconf,amsmath,graphicx}
\usepackage{amssymb}
\usepackage{hyperref}
\usepackage{caption}
\title{End-To-End Speech Recognition Using A High Rank LSTM-CTC Based Model}
\name{Yangyang Shi \qquad Mei-Yuh Hwang \qquad Xin Lei}
\address{Mobvoi AI Lab, Seattle, USA\\
  {\small \tt \{yyshi,mhwang,mikelei\}@mobvoi.com}}
\begin{document}
%
\maketitle
\begin{abstract}
Long Short Term Memory Connectionist Temporal Classification (LSTM-CTC) based end-to-end models are widely used in speech recognition due to its simplicity in training and efficiency in decoding. In conventional LSTM-CTC based models, a bottleneck projection matrix maps the hidden feature vectors obtained from LSTM to softmax output layer.  In this paper, we propose to use a high rank projection layer to replace the projection matrix. The output from the high rank projection layer is a weighted combination of vectors that are projected from the hidden feature vectors via different projection matrices and non-linear activation function. The high rank projection layer is able to improve the expressiveness of LSTM-CTC models. The experimental results show that on Wall Street Journal (WSJ) corpus and LibriSpeech data set, the proposed method achieves $4\%-6\%$ relative word error rate (WER) reduction over the baseline CTC system.
They outperform other published CTC based end-to-end (E2E) models under the condition that no external data or data augmentation is applied. Code has been made available at \url{https://github.com/mobvoi/lstm\_ctc}.
\end{abstract}
\section{Introduction}
\label{sec:intro}
 Conventional deep neural network HMM hybrid speech recognition systems \cite{Hinton2012,Xiong2017} usually require two steps in the training stage. First, a prior acoustic model such as Gaussian mixture models (GMM) is used to generate HMM state alignments for the speech training data. Based on the acoustic features and one-hot training targets generated from the state alignments, neural networks are trained to predict the frame-level state posterior probabilities. This separated two-step training process makes the acoustic model performance optimization less efficient. 

Recently, various end-to-end (E2E) models \cite{Das2018,Kim2017,Kim20171,Graves2014,Graves2006,Chiu2017,Battenberg2018,Chan2016,Sak2017,Miao2016,Sainath2017,Hannun2014} are proposed to bypass the label alignment stage to directly learn the transducer of a sequence of acoustic features to a sequence of probabilities over output tokens. These E2E systems can be categorized into CTC based models \cite{Graves2006,Graves2014,ds2,Miao2016}, sequence to sequence attention based models \cite{Chorowski_attention_2015,Chan2016,Chiu2017,Graves2012} and the combination of CTC together with sequence to sequence attention based models \cite{Zeyer2018,Kim2017,Hori2017,Das2018}.

Among the aforementioned E2E models, the CTC based models are widely investigated in the speech community \cite{Sak2015,Miao2016,Das2018,Kim20171}, due to its simplicity in training and efficiency in decoding. In CTC based models, a special blank label is introduced to identify the less informative frames. In addition, CTC based systems allow repetition of labels. In this way, CTC based models automatically infer the speech frame and label alignment (usually by a delay in time), which removes the state alignment step in training. Using highly efficient greedy decoding with no involvement of lexicon and language model, the CTC based model \cite{Sak2015} gives competitive results. In greedy decoding, the predictions are the concatenation of tokens that correspond to the spikes in posterior distribution. 

CTC loss is often used together with LSTM \cite{Sak2015,Miao2016,Graves2014}. CTC loss function imposes the conditional independence constraints for the output tokens given the whole input feature sequence. So it relies on the hidden feature vector of the current frame to make predictions. Armed with the memory mechanism, current frame's hidden feature vector from LSTM is able to capture the information from previous frames. In other words, the current frame label is not predicted based on exclusively current frame features.   

In LSTM-CTC based models, to get the posterior probability of the output labels, a projection matrix maps the hidden feature vector to the final output layer. The hidden feature vector is the output from the last layer in the multi-layer LSTMs or bidirectional LSTMs. The output layer has the same dimension as the training labels. The phonemes or the characters are usually used as labels which have smaller dimensions than the LSTM output. So the projection matrix becomes the bottleneck that limits the expressive capability of the LSTM-CTC based models. To address similar issues in language modeling, a mixture of softmaxes method \cite{Yang2017} is used to improve the performance. In this paper, we propose to use a high rank projection layer to replace the single projection matrix to improve the expressiveness of the LSTM-CTC based E2E models. 

In the high rank projection layer, one hidden feature vector is first mapped to multiple vectors by a set of projection matrices together with non-linear activation function. A weighted combination of these vectors is used as the output of the high rank projection layer. The non-linear activation function breaks the potential linear correlation among the output vectors that are obtained by mapping one feature vector via several projection matrices. So the proposed projection layer has higher rank than mapping feature vectors with one single projection matrix. 

One simple approach to decode with CTC based models is to concatenate the non-blank labels corresponding to the posterior spikes and to remove the continuously repeated output labels. However, such a simple greedy decoding method lacks the lexicon and language model information that could be leveraged to constrain the search path in decoding. In EESEN \cite{Miao2016}, a WFST based method is applied to integrate the CTC frame labels, lexicons and language models into one search graph. In this work, we follow EESEN's way of doing decoding with CTC based models.

In CTC training, the actual label sequence is obtained by inserting blank labels at the beginning, at the end and between every consistent labels in the original label sequence. The blank label has a very high prior probability. That is one reason why for the trained CTC model, the majority of frames would take blank as labels and the non-blank labels only happen in a very narrow region with peaky distribution. To address this issue, similar to EESEN \cite{Miao2016}, we apply the label distribution of the augmented label sequence used in CTC training as prior to normalize the posterior probability distribution.  

We evaluate the proposed high rank LSTM-CTC based end-to-end speech recognition on Wall Street Journal (WSJ) \cite{Paul1992} and LibriSpeech corpus \cite{Panayotov2015}. For both experiments, no external data or data augmentation is applied. On both data sets, the proposed models outperform the baseline model. For easy comparison and results reproduction, the source code for this study is released as an open source project\footnote{https://github.com/mobvoi/lstm\_ctc}.  

The rest of the paper is organized as follows. In Section 2, we briefly review the LSTM-CTC based models in the E2E speech recognition system. Then we describe the proposed high rank LSTM-CTC based models. In Section 3, we present the  experiments on WSJ and LibriSpeech benchmark data set. Finally, we give our conclusions.

\section{A High Rank LSTM-CTC Based Model}
\label{sec:main}

\subsection{LSTM-CTC}
Let $X={x_1, ..., x_T}$  denote the input sequence of acoustic feature vectors with sequence length $T$, where $x_i \in \mathbb{R}^{m}$. Given $X$, the E2E ASR system gives $p(Y|X)={p(y_{1}|X), ..., p(y_{L}|X)}$ a sequence of posterior probability vectors of the output labels, where $p(y_{i}|X)$ is a posterior probability vector of the output labels at position $i$. The dimension of the posterior probability vector $p(y_{i}|X)$ is $K$ that is the number of the target labels. The target labels usually are the phonemes or the characters. In this paper, we only use the phonemes as output labels. 

One typical problem for E2E speech recognition is that the length of output labels $L$ is often shorter than the length of input speech frames $T$. To deal with this issue in training, CTC introduces a special blank label $\phi$ that is inserted between two consecutive labels and allowing for repetitions of labels. So the label sequence $Y$ is expanded to $\Omega(Y)$ that has the same length as input sequence. To get the posterior probability of a label sequence $Y$, CTC needs to compute and sum the posterior probabilities of all the possible path $\pi$ in $\Omega(Y)$. Under the constraint that given the input sequence, the posterior probability of each label in a output sequence is conditionally independent of each other, the CTC loss is formulated as follows:

\begin{equation}
p(Y|X) = \sum_{\pi \in \Omega(Y)} p(\pi|X) = \sum_{\pi \in \Omega(Y)}\prod^{T}_{i=1}p(\pi_{i}|X).
\end{equation}

More specifically, in LSTM-CTC models, the sequence hidden feature vectors $H={h_1, ..., h_T}$ is obtained by feeding multiple layers of LSTM or bidirectional LSTM with input acoustic feature $X$. A projection matrix $M$ shared across over the whole sequence is used to map the hidden feature vectors to logit vectors of which each has $K+1$ nodes corresponding to $K+1$ labels including blank label $\phi$. The projection can be formulated as follows:
\begin{equation}
l_i = M^{t}h_i \quad i \in {1,...,T}
\label{single_projection}
\end{equation}

Softmax activation function is then applied on each logit vector $l_i$ to get the posterior probability vector $p(\pi_i)$. Normally the number of output labels $K$ is relatively small. For example, there are 71 stressed phones in WSJ data set and 43 unstressed phones in LibriSpeech data set. This projection matrix becomes the bottleneck for the expressiveness of the LSTM-CTC models. To address this issue, we proposed a high rank projection layer to replace the single projection matrix. 

\subsection{A High Rank Projection Layer}

\begin{figure}[h]
\includegraphics[width=7.5cm]{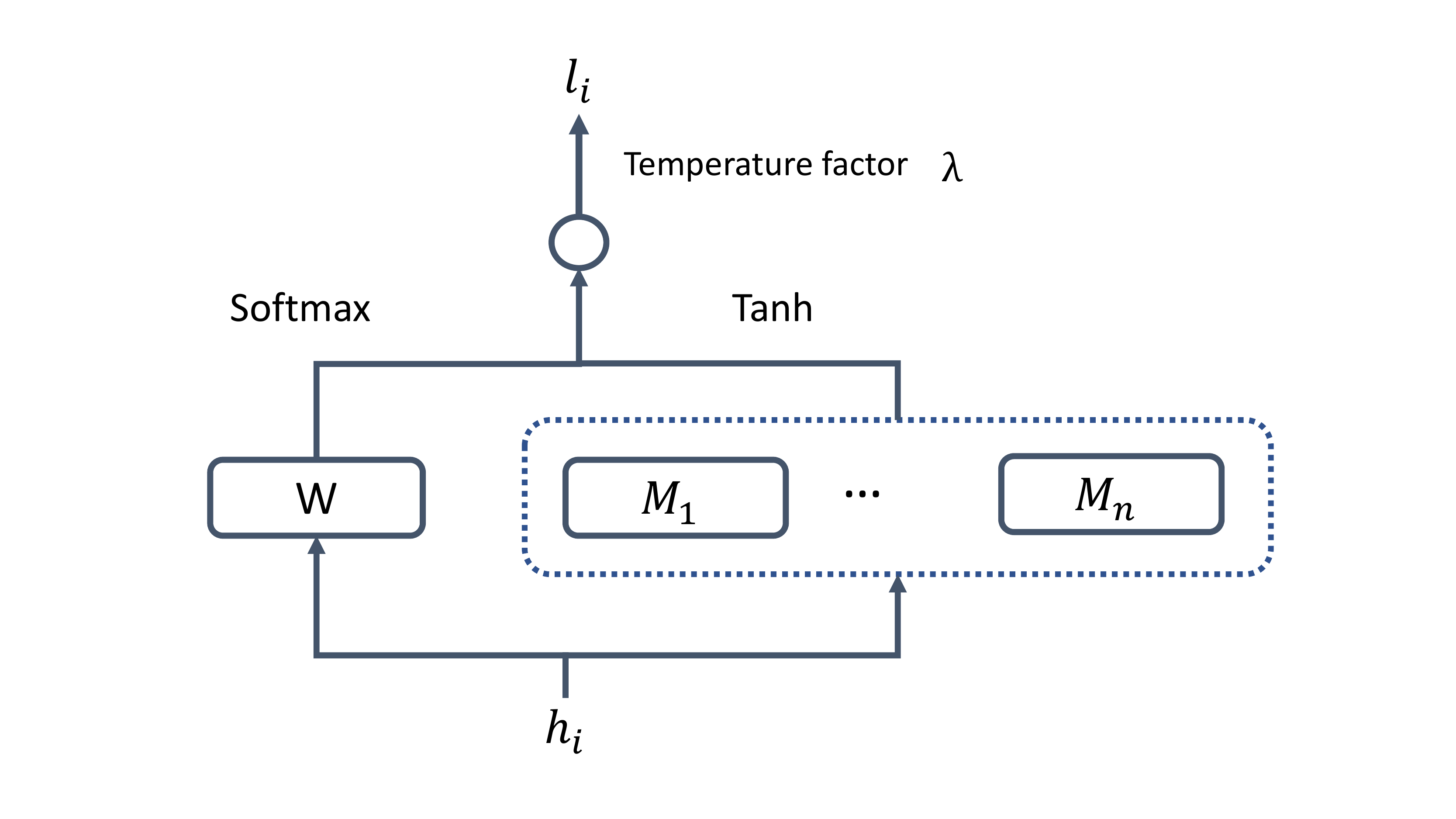}
\caption{A high rank projection layer.  A set of logit vectors are obtained by transducing the hidden feature vector $h_i$ via a set of projection matrices $M_j$ where $j\in(1, ..., n)$ and Tanh activation function. These $n$ vectors are then interpolated via a latent weight $W$. The output vector $l_i$ is obtained by scaling the weighted interpolation vector with temperature factor $\lambda$. }
\label{fig}
\vspace{-0.5pt}
\end{figure}

As illustrated in Fig.~\ref{fig}, in the high rank projection layer, a set of $n$ projection matrices are used to map the input hidden feature vector $h_i$ (of dimension $H$)  at frame $i$ to a set of logit vectors $l_{i,j}$ (each of dimension $N$).
\begin{equation}
(l_{i,1}, ..., l_{i, n}) = \tanh([M_1, ..., M_n]^{t} h_i)
\label{sets_projection}
\end{equation}
where $n$ is the predefined number of projection matrices in this layer. $[M_1, ..., M_n]$ is the concatenation of a set of projection matrices.
Each $M_j$ is of dimension $H \times N$.
The logit vector $l_i$ at speech frame $i$ is represented as an interpolation of the set of logit vectors as follows:
\begin{equation}
l_i = \lambda \sum_{j=1}^{n} w_{i,j} l_{i,j}
\label{sets_projection}
\end{equation}
where $\lambda$ is a predefined scale factor to control the smoothness of the posterior probabilities. $w_{i,j}$ is the combination weight computed at time stamp $i$ for the $j$-th logit vector. It is the softmax after mapping 
 the hidden feature vector $h_i$ to an $n$-dimensional vector via $W_{H \times n}$. 
 
\begin{equation}
\hat{w}_i  =  W^{t}h_i \label{latent}
\end{equation}

\begin{equation}
w_{i,j}  =  \frac{exp(\hat{w}_{i,j})}{\sum_{k=1}^{n} exp(\hat{w}_{i,k})}
\label{softmax}
\end{equation}

The $n$ projection matrices, $M_1, ..., M_n$, and the weight matrix $W$ are all trained jointly with the rest of network parameters. 

\subsubsection{Non-linear activation function and temperature factor}
To get a high rank projection, the non-linear activation needs to be used to break the potential linear correlation among the projection matrices in the projection layer. Without the non-linear activation, the logit vector $l_i$ at speech frame $i$ can be formulated as follows:
\begin{equation}
l_i = \sum_{j=1}^{n} w_{i,j} M_{j}^{t} h_i =  \hat{M}^{t}h_i ,
\label{moe}
\end{equation}
which is essentially the same as equation~(\ref{single_projection}). The temperature factor $\lambda$ controls the smoothness of the label output distribution. The weighted interpolation usually smooths the output probability distribution. To make the output probability distribution more discriminative, in this study, we use $\lambda \in [10,20]$ to sharp the output distribution.


\section{Experiments}
\label{sec:experiments}
\subsection{Data Sets}
 We carry out experiments on Wall Street Journal (WSJ) corpus \cite{Paul1992} and LibriSpeech corpus \cite{Panayotov2015} to verify the performance of the proposed method. The WSJ corpus is a combination of LDC93S6B and LDC94S13B data sets obtained from LDC. After data preparation, we get 81 hours of transcribed speech audio, from which $95\%$ is selected as training data, the rest $5\%$ is used as validation data. The development data (dev93) consists of 503 utterance. And the evaluation data (eval92) contains 333 utterances. LibriSpeech is an open source speech corpus\footnote{http://www.openslr.org/12/} that has almost 1000 hours read speech based on public domain audio books. Similar to WSJ data preparation, among the 960 hours' train data, we select 95$\%$ of the data for model training and the rest 5$\%$ for validation. In LibriSpeech, the development data and evaluation data are split into "clean" and "other" subsets. 

In decoding, we use WSJ provided trigram language model. In LibriSpeech experiment, to be consistent with previous studies \cite{Zhou2017}, the provided standard unpruned four-gram language model\footnote{http://www.openslr.org/resources/11/4-gram.arpa.gz} is used in decoding. 

In our experiments, the phonemes are used as CTC labels. For WSJ experiment, the CMU dictionary\footnote{http://www.speech.cs.cmu.edu/cgi-bin/cmudict}  is used as the lexicon for WFST graph building. Including the blank label, we extract 72 labels in total from CMU dictionary. In LibriSpeech experiment, we use the unstressed phonemes based lexicon\footnote{http://www.openslr.org/resources/11/librispeech-lexicon.txt} from which 44 labels are extracted as CTC labels. Due to the lack of forced alignment, CTC training can not deal with the same word with multiple pronunciations. For every word, only the first pronunciation is applied to form the lexicon. We did not use other existing models to find the best pronunciation per occurrence.

\subsection{Model Structure and Hyper-parameter Setup}
For both experiments, 120-dimensional feature vector that consists of 40-dimensional filter bank together with its first and second order derivations are calculated at each speech frame. The features are normalized via mean subtraction and variance normalization per speaker. The splice of the feature vectors from left, current and right frame (in total 360-dimensional feature vector) is used as the input to bidirectional LSTM. To speed up training, frame skipping is used. Two out of three frames are skipped during training. 
Four layers of bidirectional LSTMs are used to get the hidden feature vectors. There are 320 hidden neurons in each LSTM layer with peephole connections. The forget gate bias is set to be 5. Batch size is set to 64 for experiments on LibriSpeech and 32 for experiments on WSJ. Adam based adaptive learning rate method is used. The initial learning rate is set to 0.001 for WSJ experiments and 0.0004 for LibriSpeech experiments, respectively. The learning rate gets decayed by a factor of 0.7 for WSJ experiments and 0.5 for LibriSpeech experiments when the model does not improve over validation data. For the proposed high rank LSTM-CTC based models, we set $n$ the same as output lable size to achieve the highest rank of the projection layer. 

Due to the fact that some GPU operations are non-deterministic in tensorflow, the models trained with the same setting up multiple times would be different. For fair comparison, we use the average word error rate of five different models that are trained with the same setting up.

\subsection{Results}

Table.~\ref{tab:wsj} gives the WER comparison for different models on WSJ corpus. Comparing with our baseline model (our-LSTM-CTC), the proposed model (our-HR-LSTM-CTC) gets $6\%$ and $4\%$ relative WER reduction on dev93 and eval92, respectively.  We showed in Eq.~(\ref{moe}), our-MOM-LSTM-CTC is similar to baseline model except that it has more weight parameters. The results in Table.~\ref{tab:wsj} confirms that removing non-linear activation function and temperature factor, the simple mixture of different projection matrices does not improve over the baseline model.

\begin{table}[h]
\begin{center}
\small
\begin{tabular}{|l|c|c|c|}
  \hline method    &lm & dev93 &eval92 \\ \hline
   ESPNET\cite{espnet} &c-lstm &12.4   &8.9 \\ 
  EESEN\cite{Miao2016}&3gram  &10.9  & 7.3   \\
  CTC-PL\cite{Zhou2017}*&3gram &9.2   & 5.5   \\
   DS2\cite{ds2}*  &4gram  &5.0 &3.6 \\         \hline
  our-LSTM-CTC          &3gram &11.0   & 7.5       \\
  our-MOM-LSTM-CTC      &3gram &11.1   & 7.5       \\
  our-HR-LSTM-CTC       &3gram &10.3   &7.2        \\
\hline
\end{tabular}
\end{center}
\caption{WER($\%$) comparison for different models on WSJ dev93 and eval92 data sets. DS2 used 11940 hours audio with additional data augmentation. CTC-PL also used data augmentation. Our-LSTM-CTC is our baseline model. Our-MOM-LSTM-CTC is the mixture-of-matrices model that removes the non-linear activation function and the temperature factor in the high rank projection layer. Our-HR-LSTM-CTC is the proposed high rank LSTM-CTC model. The WER for "our-" models is the average WER of 5 models trained with the same parameter setting up.}
  \label{tab:wsj}
\end{table}

Table.~\ref{tab:librispeech} shows the WER comparison of different models on the LibriSpeech corpus. The proposed model (our-HR-LSTM-CTC) shows consistent behavior on both WSJ and LibriSpeech.

Table.~\ref{tab:wsj} and Table.~\ref{tab:librispeech} compare the results from other models using CTC loss. Due to the lack of open-sourced data, script and code, to test our models on the exact same settings as published results is difficult. To present the status of CTC loss on these two data sets, we only refer the published results here. Note some of the comparisons are not fair, as they are not trained based on the exact same data.  CTC-PL is the model trained by CTC loss together with policy learning to optimize WER. In CTC-PL, the training data is augmented through random perturbations of tempo, pitch, volume, temporal alignment, along
with adding random noise. In DS2, it uses all the public available English corpus together with data augmentation as training data. E2E-att combines sequence attention modeling together with CTC loss. It use additional 800M words for language model training. When LSTM based LM is used in decoding, E2E-att gets the state-of-the-art result on LibriSpeech. ESPNET in Table.~\ref{tab:wsj}  uses a combination of CTC loss with sequence to sequence loss. However, it does not use any effective method to leverage the language model and lexicon information in decoding. 

\begin{table}
\begin{center}
\small
\begin{tabular}{|l|c|c|c|c|c|}
  \hline method  &lm   &\multicolumn{2}{|c|}{dev} & \multicolumn{2}{|c|}{test} \\ 
                 &     & clean  & other & clean &other \\ \hline  
  CTC-PL\cite{Zhou2017}*  &4gram&5.1  & 14.3      & 5.4 & 14.7     \\
  DS2\cite{ds2}*  &4gram &-   &-           &5.3 &13.3 \\    
  E2E-att\cite{Zeyer2018}*&4gram &5.0 & 14.3      & 4.8 & 15.3     \\ 
  E2E-att\cite{Zeyer2018}*&LSTM &3.5 & 11.5      & 3.8 & 12.8     \\ \hline
  our-LSTM-CTC           &4gram &5.0 & 13.4     & 5.4 & 13.9      \\
  our-MOM-LSTM-CTC       &4gram &5.0 & 13.3     & 5.5 & 14.0      \\
  our-HR-LSTM-CTC       &4gram &4.8& 12.9 &  5.1  &13.3           \\
\hline
\end{tabular}
\end{center}
  \caption{WER($\%$) comparison for different models on LibriSpeech dev and clean data sets. DS2 is the identical system as in Table.~\ref{tab:wsj}. CTC-PL applies the same algorithm as the CTC-PL in Table.~\ref{tab:wsj}, but on LibriSpeech data with data augmentation. E2E-att used external data for language model training. Our-LSTM-CTC is our baseline CTC model trained on LibriSpeech. Our-MOM-LSTM-CTC is the model that removes the non-linear activation function and the temperature factor. Our-HR-LSTM-CTC is the proposed high rank LSTM-CTC model. The WER for "our-" models is the average WER of 5 models trained with the same parameter setting up. } 
  \label{tab:librispeech}
\end{table}

\section{Conclusions}
In this paper, a high rank projection layer is proposed to replace the bottleneck projection matrix in conventional LSTM-CTC based models for E2E speech recognition. The output of the high rank projection layer is a weighted combination of multiple vectors that are obtained by feeding the hidden feature vector to a set of projection matrices and going through a non-linear activation function. On two benchmark corpora, WSJ and LibriSpeech, the proposed high rank LSTM-CTC model outperformed the baseline CTC model. On WSJ corpus, compared with baseline model, the proposed model got nearly $6\%$ relative WER reduction on dev93 and $4\%$  reduction on eval92. On LibriSpeech corpus, the proposed model improved the baseline model by relative WER reduction  $6\%$ on test-clean and $4\%$ on test-other, dev-clean and dev-other subsets.

\vfill\pagebreak

\bibliographystyle{IEEEbib}
\bibliography{icassp2019}

\end{document}